\documentclass[conference]{IEEEtran}
\IEEEoverridecommandlockouts

\usepackage{cite}
\usepackage{amsmath,amssymb,amsfonts}
\usepackage{graphicx}
\usepackage{textcomp}
\usepackage{xcolor}
\usepackage{multirow}%
\usepackage{gensymb}
\usepackage{hyperref}
\usepackage{booktabs}
\usepackage{threeparttable}
\usepackage{float}
\usepackage{manyfoot}%
\usepackage{algorithm}%
\usepackage{algorithmicx}%
\usepackage{algpseudocode}%
\usepackage{orcidlink}
\usepackage{subcaption}
\def\BibTeX{{\rm B\kern-.05em{\sc i\kern-.025em b}\kern-.08em
    T\kern-.1667em\lower.7ex\hbox{E}\kern-.125emX}}
\begin{document}

\title{Self-Supervised Graph Representation Learning for In-The-Wild Wearable and Smartphone based Emotion Recognition \\

\thanks{Code at: \url{https://github.com/GiannisZgs/Wearable_ER_Graph_SSL}. \\
This work is supported by Khalifa University of Science and Technology.
\copyright~2025 IEEE.  Personal use of this material is permitted. Permission from IEEE must be obtained for all other uses, in any current or future media, including reprinting/republishing this material for advertising or promotional purposes, creating new collective works, for resale or redistribution to servers or lists, or reuse of any copyrighted component of this work in other works. DOI: 10.1109/ICASSP49660.2025.10888648}
}

\author{\IEEEauthorblockN{Ioannis Ziogas}
\IEEEauthorblockA{\textit{Dpt. of Biomedical Eng. and Biotechnology} \\
\textit{Khalifa University of Science and Technology}\\
Abu Dhabi, UAE\\
ioannis.ziogas@ku.ac.ae \orcidlink{https://orcid.org/0000-0003-0615-322X}}
\and
\IEEEauthorblockN{Leontios J. Hadjileontiadis}
\IEEEauthorblockA{\textit{Dpt. of Biomedical Eng. and Biotechnology} \\
\textit{Khalifa University of Science and Technology} \\ 
Abu Dhabi, UAE\\
\textit{Dpt. of Electrical and Computer Eng.}\\
\textit{Aristotle University of Thessaloniki} \\ 
Thessaloniki, Greece\\
leontios.hadjileontiadis@ku.ac.ae}
\and
\IEEEauthorblockN{Ahsan H. Khandoker}
\IEEEauthorblockA{\textit{Dpt. of Biomedical Eng. and Biotechnology} \\
\textit{Khalifa University of Science and Technology}\\
Abu Dhabi, UAE\\
ahsan.khandoker@ku.ac.ae}
\and
\IEEEauthorblockN{Aamna Al Shehhi}
\IEEEauthorblockA{\textit{Dpt. of Biomedical Eng. and Biotechnology} \\
\textit{Khalifa University of Science and Technology}\\
Abu Dhabi, UAE\\
aamna.alshehhi@ku.ac.ae}
}

\maketitle

\begin{abstract}
Wearable and smartphone-based emotion recognition (WER) remains a challenging setting in affective computing, due to the notorious difficulty and bias associated with in-the-wild label collection. The high inter-and intra-subject emotional variability motivates us to explore WER modeling through graph node classification in a limited resources learning scheme powered by Self-Supervised Learning (SSL) graph masking augmentation tasks. We employ a subgraph sampling approach during training, utilizing labeled and unlabeled data, along with supervised, semi-supervised, and SSL mechanisms in a multi-task inductive graph neural network architecture. Our evaluations on K-EmoPhone through leave-one-group-out cross-validation in the binary arousal and valence tasks yield average accuracy gains of \textbf{4.3\%} and \textbf{7.8\%}, compared to the full resource setting, utilizing only \textbf{20\%} and \textbf{25\%} of the labels, respectively. Our model analysis sheds light on the relation of SSL graph augmentations to emotional arousal and valence and justifies the approach of SSL-driven subgraph training for in-the-wild WER. 
\end{abstract}


\section{Introduction}
Accurate analysis of emotional manifestations that unfold during typical daily life interactions and experiences, has long been a challenging scenario for affective computing systems \cite{kang2023kemophone}. In part, the advent of ubiquitous and unobtrusive data collection through smartphones and wearable devices along with Artificial Intelligence (AI)-powered computing, have addressed this challenge by utilizing the scarce 
affect labels for emotion recognition (ER) \cite{schuller2018}. 
Hence, in-the-wild ER presents a challenging landscape, increasingly relevant in the discussion of \textit{how to efficiently exploit the limited available resources}, in terms of affect labels.

Concurrently, Self-Supervised Learning (SSL) is a novel paradigm in AI that does not require labeled instances to learn \cite{balestriero2023}. Pre-text augmentation tasks, 
facilitate 
SSL by fostering a generation of self-supervisory signals from the data, through perturbations of the data structure \cite{spathis2022}. Computer vision-inspired augmentations 
have been applied to time series \cite{dissanayake2022a,wu2023sslWearableEmo}, yet they do not possess the capacity to model 
temporal dependencies, temporal ordering and the high-dimensional nature of univariate and multivariate signals \cite{zhang2023sslreview}. In wearable ER, SSL approaches are limited in number \cite{Dissanayake2022SigRep:Learning,wu2023}; whereas, GNNs have been sparsely used for SSL in wearable affective computing \cite{Yang2023APhysiology,Yang2020APhysiology}. 


SSL on Graph Neural Networks (GNNs), on the other hand, faces significant challenges in effectively learning informative representations, as it has to capture both the local and global topology of the graph, as well as the node or edge attributes' information \cite{xie2023}. Graph-specific augmentations, such as node masking and edge dropping, aim to perturb the structure of the adjacency matrix of the graph, altering the graph structure and connectivity \cite{you2020graphCL}. 
To that end, the representation of the time series data as a graph structure provides a novel perspective on the inter-individual emotion analysis problem by connecting emotionally loaded data samples across multiple biosignal and smartphone modalities. Unlike CNN and Transformer-based approaches, the graph structure explicitly embeds labeled and unlabeled instances in a structural similarity space.   

\begin{figure*}[ht!]
    \centering
    \includegraphics[width=0.75\textwidth,height = 5cm]{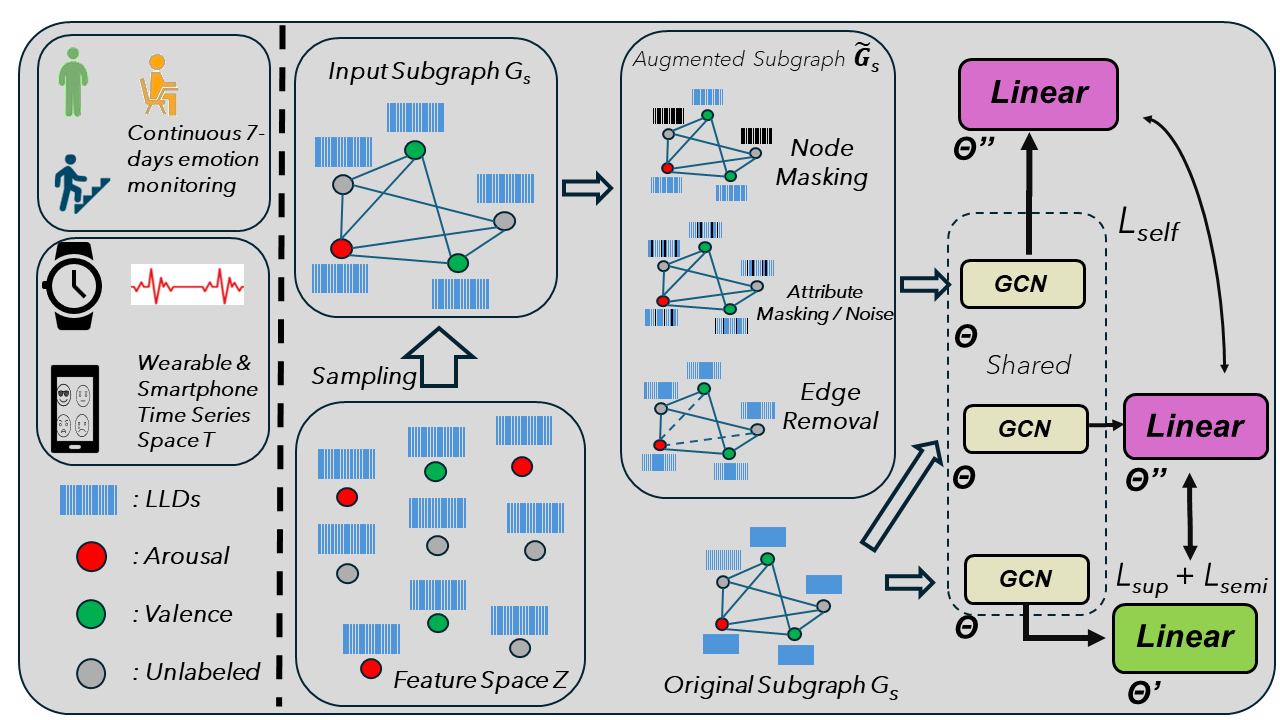}
    \caption[]{\footnotesize{\textsf{Our proposed SSL-based graph representation learning framework.}}}
    \label{fig1}
\end{figure*}

In this work, we propose an SSL-based graph representation learning approach to in-the-wild ER through wearable and smartphone data, in a limited resource setting, where only a few labels are available. 
Motivated by the success of a similar approach in audio event modeling \cite{shirian2022}, we formulate the ER problem as an \textit{inductive} graph node classification task, where we investigate the potential of SSL in enhancing performance 
by training on relatively small subgraphs. Our experimental results indicate that the construction of smaller graphs ameliorates the influence of unlabeled instances, compared to using a single large graph. We demonstrate how performance increases by gradually integrating additional knowledge into our system, in the form of \textit{graph masking SSL tasks}. 
Our results on the K-EmoPhone database \cite{kang2023kemophone} validate our hypothesis that limited resource settings benefit from SSL techniques, demonstrating substantial increases in performance over graph baselines on the full resource scenario. 


\section{Proposed Method}
\label{section2}


\subsection{Graph Construction}
\label{graphConstruct}
We consider a feature encoder $f: T \rightarrow Z$ that performs a mapping from the input space $T$ of time series data, to an embedding space $Z$. Through a sample-wise approach, time series segments in $T$ correspond to embeddings in $Z$, and each segment is modeled as a vertex/node $V$ in a graph structure, where embeddings $Z$ serve as the node attributes.
We argue that inter-subject similarities and differences can be captured through a sufficiently small graph, and hence our subgraph sampling approach during training is inspired by the proposed framework of Shirian \textit{et al.} \cite{shirian2022}. Specifically, given an initial set of time series segments $S$, with and without labels, we select a labeled subset $L \subseteq S$ and an unlabeled subset $U \subseteq S$, so as that $S = L \cup U$ holds. 

We first connect the labeled nodes to an undirected graph $G = (V,E)$, by establishing connections between nodes based on a similarity metric criterion. Specifically, each node $u_i \in L$ is connected through edges $e_{ij}$ with weight $a_{ij} = 1$ to its $k$-nearest neighbors in $L$, given that nodes $u_i, u_j$ have the same label. 
Further, each node $u_i in L$ is connected to its $m$-farthest neighbors in $L$, irrespective of the label, through edges $e_{ij}$ with an edge weight of $a_{ij} = -1$. 
For the unlabeled nodes,
we connect every node $u_i \in U$ to its $k$-nearest and $m$-farthest neighbors in $S$, with weighted edges $a_{ij} = 1$ and $-1$, respectively. The above process yields an adjacency matrix $A$ for the undirected graph $G$, where $A = \{a_{ij}\}^{|S|}$.
During training, a fixed percentage of labeled $L_s \in L$ and unlabeled $U_s \in U$ nodes is randomly sampled from $S$ \cite{zeng2019graphSAINT}, while maintaining subject integrity and class balance on the sampled segments, and at each epoch, a single subgraph $G_s$ is constructed from $(L_s,U_s)$ as described above.

\subsection{Multi-Task Training with Limited Labels}
We adopt a multi-task learning approach to 
exploit the supervision from the labeled part of the data, while leveraging self-supervisory signals from the unlabeled instances (see Fig.\ref{fig1}). In this vein, a graph encoder $g_{\theta}: Z \rightarrow H$ with parameters $\theta$, is trained on the input graph embeddings $Z$ from the feature encoder $f$, and outputs latent space embeddings $H$. The graph encoder is shared between two submodules; the first submodule $g_{\theta'}$ with parameters $\theta'$, performs semi-supervised node classification by using the labels and an SSL pseudo-labeling task, while the second submodule $g_{\theta''}$ with parameters $\theta''$, performs SSL reconstruction tasks in the embedding space. 
Our model, therefore, tries to solve an optimization problem that involves the simultaneous learning of parameters $\theta, \theta', \theta''$, as formulated below: 

\begin{equation}
\label{totalLoss}
\begin{split}
    \underset{\theta, \theta', \theta''}{\operatorname{min}} &[\mathcal{L}_{sup}(\theta,\theta',G_s) + \lambda_1 \mathcal{L}_{semi}(\theta,\theta',G_s) \\ &+ \lambda_2 \mathcal{L}_{self}(\theta,\theta'',G_s,G_s')]
\end{split}
\end{equation}

where $\lambda_1$ and $\lambda_2$ are the regularization parameters for the semi- and self-supervised losses, respectively.

\subsubsection{Supervised Node Classification Task}
For the labeled nodes $L_s$ with labels $y_s$ of the subgraph $G_s$, we compute the cross-entropy loss over the softmax outputs $\hat{y_s}$ of the emotion class logits: 

\begin{equation}
    \label{CELoss_sup}
    \mathcal{L}_{sup} = - \sum_{u_i \in L_s}{y_s \log{\hat{y_s}}} 
\end{equation}

\subsubsection{Semi-Supervised Pseudolabeling Task}
For the unlabeled set of nodes $U_s$, we employ the network predictions $\hat{y_s}$ to formulate an additional regularization term, that penalizes the prediction uncertainty of the network. Since the subgraph consists of both labeled and unlabeled nodes, the network has labeled information that relates to the labeled connections of unlabeled nodes. Based on that, we judge the confidence of the network in producing an output distribution that is characterized by low entropy values. We formulate this task through the below loss:

\begin{equation}
    \label{CELoss_semi}
    \mathcal{L}_{semi} = - \sum_{u_i \in U_s}{\hat{y_s} \log{\hat{y_s}}} 
\end{equation}

This procedure yields \textit{pseudolabels} $\hat{y_s}$ at each training iteration; we store these pseudolabels and use them as additional supervision during inference (see subsection \ref{inference}). 

\subsubsection{Self-Supervised Reconstruction Tasks}
To exploit structural information from the sampled subgraphs at each epoch, 
we produce a perturbed version of the input subgraph $\Hat{G_s}$, and then attempt to bring closer the embeddings $W$ and $\Hat{W}$ of the original $G_s$ and the perturbed $\Hat{G_s}$, 
calculating an SSL loss over a set of nodes $V$, as follows:

\begin{equation}
    \mathcal{L}_{self} = \frac{1}{|V|} \|\Hat{W} - W\|^{2}
\end{equation}

We employ four commonly used augmentations found in the graph SSL literature\cite{you2020graphCL,zhu2021pyGCL,shirian2022}: 
\textbf{\textit{a) Node Masking - }} each node is masked with a probability $p_m$ that follows an i.i.d. uniform distribution, 
\textbf{\textit{b) Node Attribute Masking - }} node attributes are masked with a probability $p_m$ that follows an i.i.d. uniform distribution, 
\textbf{\textit{c) Gaussian Noise Addition - }} we produce a noisy graph $\Hat{G_s}$ with node attributes $\hat{z_i} = z_i + x_i$, where $x_i \sim \mathcal{N}(0,1)$, by adding Gaussian noise 
with probability $p_n$ that follows an i.i.d. uniform distribution,  
\textbf{\textit{d) Edge Removal - }} we perturb the graph connectivity by randomly removing edges from the graph structure with a probability $p_r$ following an i.i.d. uniform distribution. 
For all tasks except node masking, the SSL loss is calculated over the whole collection of subgraph nodes $V_s: (L_s,U_s)$, whereas for the node masking task only the set of nodes that have been masked $V_m: (L_m,U_m)$ is used.

\subsection{Inference}
\label{inference}
Since our approach is an inductive graph learning approach, at inference, 
we sample a subset of labeled and unlabeled nodes from the unseen data, to form an inference subgraph $G_{is}$, as was done during training; for the unlabeled nodes, we assign to each node the most commonly appearing pseudo label for each instance, calculated throughout the multiple training iterations from the semi-supervised loss. 

\section{Experiments}
\label{section3}

\subsection{K-EmoPhone Database}
\label{data}
We utilize the publicly available K-EmoPhone dataset  \cite{kang2023kemophone}, 
that contains data collected from 77 participants over a 7-day period. Data include affective reports through the experience sampling method (ESM), pre- and post-study surveys, demographics, smartphone modalities related to the connectivity, call, and internet logs, battery status, human activity, and location, to name a few, as well as physiological indicators (accelerometer, electrodermal activity, skin temperature, heart rate), with a total of 5589 ESM responses that serve as affective labels. For the current analysis, we 
keep a subset of 2619 ESM responses (47 participants) that fully complied with the study guidelines. 
We use the arousal (A) and valence (V) annotations, by binarizing the 7-point $\{-3,+3\}$ scale ratings of A and V to obtain 1586/1063 Low $(-3,0)$ and 1033/1556 High (above 0) A/V ratings, respectively.  

    

\subsection{Data Processing}
\label{procFeatEx}
We implement the data processing procedure followed by Kang \textit{et al.} \cite{kang2023kemophone}; we consider a window-wise analysis, where each ESM response denotes the end of a segment. We minimally pre-process the wearable sensors' and smartphone data, by resampling recordings and segment-wise normalization. Since the duration of a felt emotional state is not known \cite{kang2023kemophone}, we consider multiple window durations that span from 30 seconds to 6 hours in the past and extract low level descriptors (LLDs) for each of these segments. The final feature vector for each ESM response is of size 3356. We perform feature selection according to \cite{kang2023kemophone}, on each subgraph by using only labeled training nodes, to reduce the feature vector dimensionality to $\approx10\%$ of its initial size. 


\subsection{Evaluation and Training Settings}
We evaluate our proposed in-the-wild ER subgraph inductive learning approach through a  
Leave-One-Group-Out (LOGO) cross-validation scheme, i.e. a stratified w.r.t. the emotion labels 43/3/1 subject(s) train/dev/test split.
For the subgraph construction during training, at each LOGO iteration, we randomly consider $\approx 20\%$ or $L = 11$ participants of the training data as labeled, while the remaining training set $U = 32$ is considered unlabeled. 
For arousal, we sample $L_s =11, U_s = 6$ subjects, whereas for valence, we sample $L_s = 9, U_s = 5$ subjects from the total training set and connect them to the subgraph according to \ref{graphConstruct} with $k=2$ nearest and $m=1$ farthest neighbors. 

\begin{table*}[ht!]
    \centering
    \caption{Leave-one-group-out CV results for the binary Arousal/Valence (A/V) classification tasks - 25\% / 20\% of the labels, and 11 / 9 subjects are used for A/V, respectively}
    \begin{tabular}{c c c c | c c c c } 
    \toprule
    
    \multicolumn{4}{c|}{\textbf{GCN Model objective}}& \multicolumn{1}{c}{\textbf{A/V Acc. (SD)}}    
    & \multicolumn{1}{c}{\textbf{A/V F1 (SD)}} & \multicolumn{1}{c}{\textbf{Low A/V F1 (SD)}} & \multicolumn{1}{c}{\textbf{High A/V F1 (SD)}} \\
    \midrule 
    

    
    \multicolumn{4}{c|}{100\% labels (fully sup.)} & 0.541/0.513 (0.122/0/150) &\textbf{0.478}/0.436 (0.097/0.112) & 0.552/\textbf{0.353} (0.192/0.195) & \textbf{0.405}/0.519 (0.192/0.195)\\ 
    \multicolumn{4}{c|}{25\% (A) / 20\% (V) labels + unlabeled} & 0.549/0.572 (0.147/0/148) & 0.458/0/455 (0.109/0.092) & 0.627/0.311 (0.181/0/204) & 0.289/0.598 (0.181/0/204) \\

    \midrule
    
    \multicolumn{4}{c|}{pseudolabeling (semi-sup.)} & 0.574/0.554 (0.136/0.165) & 0.463/0.423 (0.091/0.095) & 0.623/0.314 (0.217/0.252) & 0.303/0.531 (0.217/0.252) \\ 
    
    \midrule
    
    \multicolumn{4}{c|}{node attribute noise} & 0.559/0.586 (0.118/0.145) & 0.463/0.453 (0.074/0.075) & 0.594/0.298 (0.200/0.234) & 0.333/0.607 (0.200/0.234) \\
    \multicolumn{4}{c|}{node masking} & 0.575/\textbf{0.591} (0.133/0.148) & 0.465/0.460 (0.083/0.082) & 0.624/0.311 (0.181/0.234) & 0.306/0.609 (0.181/0.234) \\
    \multicolumn{4}{c|}{node attribute masking} & 0.581/0.589 (0.133/0.159) & 0.462/\textbf{0.464} (0.078/0.105) & 0.631/0.309 (0.202/0.240) & 0.293/\textbf{0.620} (0.202/0.240) \\
    \multicolumn{4}{c|}{edge removal} & 0.563/0.574 (0.145/0.143) & 0.458/0.454 (0.091/0.084) & 0.625/0.316 (0.188/0.216) & 0.291/0.592 (0.188/0.216) \\
    
    \midrule
    
    \multicolumn{4}{c|}{node attribute noise + pseudolabeling} & 0.580/0.558 (0.157/0.196) & 0.461/0.415 (0.091/0.113) & 0.621/0.281 (0.224/0.249) & 0.300/0.550 (0.224/0.249) \\
    \multicolumn{4}{c|}{node masking + pseudolabeling} & \textbf{0.584}/0.557 (0.142/0.171) & 0.454/0.434 (0.103/0.099) & \textbf{0.644}/0.276 (0.201/0.220) & 0.263/0.592 (0.201/0.220) \\
    \multicolumn{4}{c|}{node attribute masking + pseudolabeling} & 0.553/0.563 (0.146/0.180) & 0.438/0.426 (0.101/0.119) & 0.579/0.297 (0.244/0.254) & 0.297/0.555 (0.244/0.254) \\
    \multicolumn{4}{c|}{edge removal + pseudolabeling} & 0.576/0.564 (0.143/0.187) & 0.450/0.415 (0.098/0.115) & 0.635/0.268 (0.212/0.254) & 0.264/0.563 (0.212/0.254) \\
   
    \bottomrule    
    \end{tabular}%
    \label{results_table}
\end{table*}

\begin{figure*}[ht!]
\begin{subfigure}{0.33\textwidth}
\centering
\includegraphics[width=0.95\textwidth,height=3.25cm]{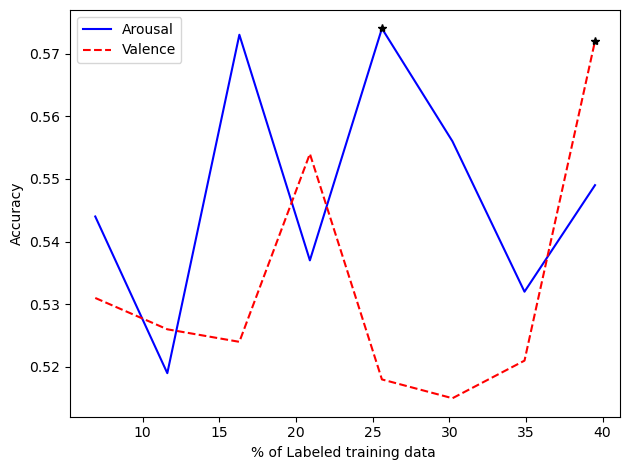}
\label{fig2a}
\end{subfigure}%
\begin{subfigure}{0.33\textwidth}
\centering
\includegraphics[width=0.95\textwidth,height=3.25cm]{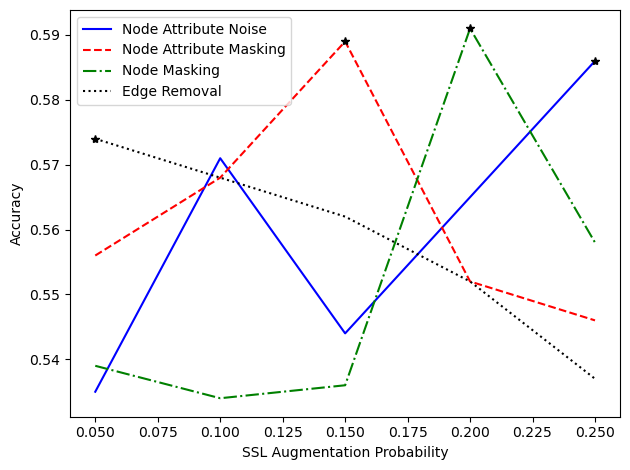}
\label{fig2b}
\end{subfigure}%
\begin{subfigure}{0.33\textwidth}
\centering
\includegraphics[width=0.95\textwidth,height=3.25cm]{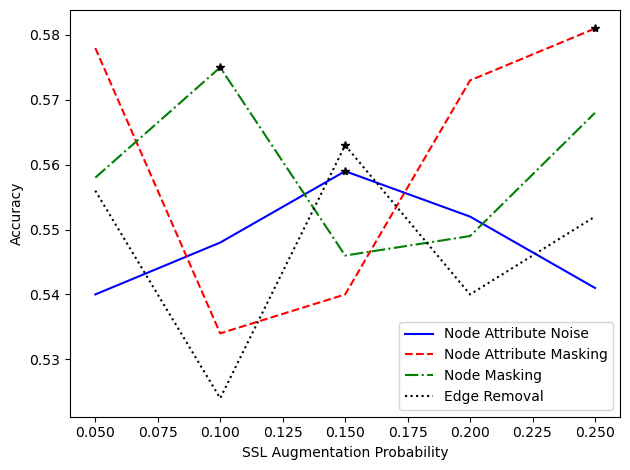}
\label{fig2c}
\end{subfigure}%
\caption[]{\footnotesize{\textsf{Impact on accuracy (y-axis) (a): percentage of labeled training data (x-axis) for arousal and valence in the without-SSL semi-supervised setting, (b,c): selection probability of an SSL task (x-axis), for (b) valence / (c) arousal, and 20\% / 25\% labeled data size, respectively.}}}
\label{fig2}
\end{figure*}

The shared graph encoder $g_\theta$ is modeled as a Graph Convolution Network (GCN) \cite{kipf2016} with 3 graph convolution layers with hyperbolic tangent activations and a hidden size of 96. The projectors $g_{\theta'}$ and $g_{\theta''}$ are implemented as simple one-layered linear heads without activation. In the case of $g_{\theta''}$ an output embedding size of 96 is selected. 
A dropout rate of $50\%$ \cite{srivastava2014dropout} and a label smoothing rate of $10\%$ \cite{pereyra2017labelsmoothing} were used. The selection of all model hyperparameters was based only on validation data. We train all networks using a constant learning rate of $0.0055$ with the Adam optimizer \cite{kingma2014Adam} with a weight decay of 1e-5, and keep the hyperparameters that yield the best validation loss. We choose the regularization parameters $\lambda_1 = 0.3$ and $\lambda_2 = 0.2$. All networks were implemented with PyTorch on a single NVIDIA RTX6000 ADA GPU. 

\section{Results}
\label{section4}

Table \ref{results_table} reports the average and standard deviation of our four evaluation metrics across the LOGO validation, for all the variants of our subgraph SSL approach. We observe that our proposed limited resources settings are performing on par or even better than the full resource setting. Compared to the fully supervised setting, increases of \textbf{4.3\%} in accuracy with the node masking and pseudolabeling SSL tasks for arousal, and \textbf{7.8\%} in accuracy and \textbf{3\%} increase in macro-F1 with the node masking and node attribute masking tasks, are observed. Moreover, the utilized labels have been reduced to a mere \textbf{25\%} for arousal and \textbf{20\%} for valence, in contrast to previous efforts that used \textbf{100\%} of the labels \cite{kang2023kemophone}. In general, we observe a gradual increase in performance by moving from the fully supervised to the lightly supervised setting with strong SSL regularization, as was expected by our initial hypothesis. We should note that in both cases, the fully supervised baseline performs better in predicting instances of the minority high arousal and low valence class.

\subsection{Ablation Studies}
\subsubsection{Effect of Labeled Data Size}
In Fig. \ref{fig2}a, we investigate labeled data percentage as a defining parameter of our system's performance. 
The results of this investigation for the semi-supervised setting, demonstrate a considerable variance in the system performance; more labeled data does not always correspond to higher accuracy. This also explains the superiority of our subgraph sampling SSL approach to the fully supervised setting; a larger graph with more labels may not always be beneficial for heterogeneous tasks such as in-the-wild ER. 

\subsubsection{Effect of SSL Graph Augmentation Probability}
We also investigate the strength of the utilized SSL augmentation tasks in the SSL setting without the pseudo labeling task in terms of the probability of perturbing nodes, edges, or node attributes in our subgraphs. In Fig.\ref{fig2}b, for the valence case, there is a gradual increase of performance until relatively higher probability values for all tasks, except edge removal. This may be attributed to the strength of edge removal as a structural perturbation on the graph; this result indicates the significance of the graph construction technique in connecting similar and dissimilar emotions. 
Conversely, graph augmentations that affect node information seem to have a milder effect, implying that emotional valence information that relates to inter- and intra-subject differences is encoded in the connectivity of the graph, rather than the segments' features.
In the case of arousal (Fig.\ref{fig2}c), 
node masking and noise addition seem to affect the performance more than edge removal 
This behavior may indicate a greater dependency of arousal prediction on the node attributes, rather than the connectivity of the graph.

\section{Conclusion}
\label{section5}
Our work contributes a novel perspective on in-the-wild WER with limited labels through wearable and smartphone time series data, by examining an SSL graph node classification approach on the foundation of subgraph construction through sampling. Through comprehensive evaluations and ablations on the 7 days-long K-EmoPhone dataset, we validate the theoretical advantages of our subgraph SSL approach, instead of a single large graph for low-resource WER. Our method does not require modality-specific encoders, and performs well under limited resources in terms of labels and data, thus holding great promise for WER in longer continuous monitoring periods and more diverse populations. 




{
\bibliography{mendeley_refs,ssl_refs,affect_refs,sysRev_refs}}

@article{balestriero2023,
    title = {{A Cookbook of Self-Supervised Learning}},
    year = {2023},
    journal = {arXiv preprint arXiv:2304.12210 (2023)},
    author = {Balestriero, Randall and Ibrahim, Mark and Sobal, Vlad and Morcos, Ari and Shekhar, Shashank and Goldstein, Tom and Bordes, Florian and Bardes, Adrien and Mialon, Gregoire and Tian, Yuandong and Schwarzschild, Avi and Wilson, Andrew Gordon and Geiping, Jonas and Garrido, Quentin and Fernandez, Pierre and Bar, Amir and Pirsiavash, Hamed and LeCun, Yann and Goldblum, Micah},
    month = {4},
    url = {https://arxiv.org/abs/2304.12210v1},
    arxivId = {2304.12210}
}

@article{kingma2014Adam,
    title = {{Adam: A Method for Stochastic Optimization}},
    year = {2014},
    journal = {3rd International Conference on Learning Representations, ICLR 2015 - Conference Track Proceedings},
    author = {Kingma, Diederik P. and Ba, Jimmy Lei},
    month = {12},
    publisher = {International Conference on Learning Representations, ICLR},
    url = {https://arxiv.org/abs/1412.6980v9},
    arxivId = {1412.6980}
}

@article{zhu2021pyGCL,
    title = {{An Empirical Study of Graph Contrastive Learning}},
    year = {2021},
    author = {Zhu, Yanqiao and Xu, Yichen and Liu, Qiang and Wu, Shu},
    journal = {arXiv preprint arXiv:2109.01116v2},
    month = {9},
    url = {https://arxiv.org/abs/2109.01116v2},
    arxivId = {2109.01116}
}

@article{spathis2022,
    title = {{Breaking away from labels: The promise of self-supervised machine learning in intelligent health}},
    year = {2022},
    journal = {Patterns},
    author = {Spathis, Dimitris and Perez-Pozuelo, Ignacio and Marques-Fernandez, Laia and Mascolo, Cecilia},
    number = {2},
    month = {2},
    pages = {100410},
    volume = {3},
    publisher = {Elsevier},
    doi = {10.1016/J.PATTER.2021.100410},
    issn = {2666-3899}
}

@article{srivastava2014dropout,
    title = {{Dropout: A Simple Way to Prevent Neural Networks from Overfitting}},
    year = {2014},
    journal = {Journal of Machine Learning Research},
    author = {Srivastava, Nitish and Hinton, Geoffrey and Krizhevsky, Alex and Sutskever, Ilya and Salakhutdinov, Ruslan},
    number = {56},
    pages = {1929--1958},
    volume = {15},
    url = {http://jmlr.org/papers/v15/srivastava14a.html},
    issn = {1533-7928}
}

@article{Yang2023APhysiology,
    title = {{A Media-Guided Attentive Graphical Network for Personality Recognition Using Physiology}},
    year = {2023},
    journal = {IEEE Transactions on Affective Computing},
    author = {Yang, Hao Chun and Lee, Chi Chun},
    number = {2},
    month = {4},
    pages = {931--943},
    volume = {14},
    publisher = {Institute of Electrical and Electronics Engineers Inc.},
    doi = {10.1109/TAFFC.2021.3090040},
    issn = {19493045}
}

@article{Yang2020APhysiology,
    title = {{A Siamese Content-Attentive Graph Convolutional Network for Personality Recognition Using Physiology}},
    year = {2020},
    journal = {ICASSP, IEEE International Conference on Acoustics, Speech and Signal Processing - Proceedings},
    author = {Yang, Hao Chun and Lee, Chi Chun},
    month = {5},
    pages = {4362--4366},
    volume = {2020-May},
    publisher = {Institute of Electrical and Electronics Engineers Inc.},
    isbn = {9781509066315},
    doi = {10.1109/ICASSP40776.2020.9054226},
    issn = {15206149}
}

@article{you2020graphCL,
    title = {{Graph Contrastive Learning with Augmentations}},
    year = {2020},
    author = {You, Yuning and Chen, Tianlong and Sui, Yongduo and Chen, Ting and Wang, Zhangyang and Shen, Yang},
    journal = {NIPS'20: Proceedings of the 34th International Conference on Neural Information Processing Systems},
    number = {488},
    month = {10},
    pages = {5812 - 5823},
    url = {https://arxiv.org/abs/2010.13902v3},
    arxivId = {2010.13902}
}

@article{zeng2019graphSAINT,
    title = {{GraphSAINT: Graph Sampling Based Inductive Learning Method}},
    year = {2019},
    journal = {8th International Conference on Learning Representations, ICLR 2020},
    author = {Zeng, Hanqing and Zhou, Hongkuan and Srivastava, Ajitesh and Kannan, Rajgopal and Prasanna, Viktor},
    month = {7},
    publisher = {International Conference on Learning Representations, ICLR},
    url = {https://arxiv.org/abs/1907.04931v4},
    arxivId = {1907.04931}
}

@article{kang2023kemophone,
    title = {{K-EmoPhone: A Mobile and Wearable Dataset with In-Situ Emotion, Stress, and Attention Labels}},
    year = {2023},
    journal = {Scientific Data 2023 10:1},
    author = {Kang, Soowon and Choi, Woohyeok and Park, Cheul Young and Cha, Narae and Kim, Auk and Khandoker, Ahsan Habib and Hadjileontiadis, Leontios and Kim, Heepyung and Jeong, Yong and Lee, Uichin},
    number = {1},
    month = {6},
    pages = {1--21},
    volume = {10},
    publisher = {Nature Publishing Group},
    url = {https://www.nature.com/articles/s41597-023-02248-2},
    doi = {10.1038/s41597-023-02248-2},
    issn = {2052-4463},
    pmid = {37268686}
}

@article{pereyra2017labelsmoothing,
    title = {{Regularizing Neural Networks by Penalizing Confident Output Distributions}},
    year = {2017},
    journal = {5th International Conference on Learning Representations, ICLR 2017 - Workshop Track Proceedings},
    author = {Pereyra, Gabriel and Tucker, George and Chorowski, Jan and Kaiser, Łukasz and Hinton, Geoffrey},
    month = {1},
    publisher = {International Conference on Learning Representations, ICLR},
    url = {https://arxiv.org/abs/1701.06548v1},
    arxivId = {1701.06548}
}

@article{shirian2022,
    title = {{Self-Supervised Graphs for Audio Representation Learning With Limited Labeled Data}},
    year = {2022},
    journal = {IEEE Journal on Selected Topics in Signal Processing},
    author = {Shirian, Amir and Somandepalli, Krishna and Guha, Tanaya},
    number = {6},
    month = {10},
    pages = {1391--1401},
    volume = {16},
    publisher = {Institute of Electrical and Electronics Engineers Inc.},
    doi = {10.1109/JSTSP.2022.3190083},
    issn = {19410484},
    arxivId = {2202.00097}
}

@article{xie2023,
    title = {{Self-Supervised Learning of Graph Neural Networks: A Unified Review}},
    year = {2023},
    journal = {IEEE Transactions on Pattern Analysis and Machine Intelligence},
    author = {Xie, Yaochen and Xu, Zhao and Zhang, Jingtun and Wang, Zhengyang and Ji, Shuiwang},
    number = {2},
    month = {2},
    pages = {2412--2429},
    volume = {45},
    publisher = {IEEE Computer Society},
    doi = {10.1109/TPAMI.2022.3170559},
    issn = {19393539},
    pmid = {35476575},
    arxivId = {2102.10757}
}

@article{dissanayake2022a,
    title = {{Self-supervised Representation Fusion for Speech and Wearable Based Emotion Recognition}},
    year = {2022},
    journal = {Proceedings of the Annual Conference of the International Speech Communication Association, INTERSPEECH},
    author = {Dissanayake, Vipula and Seneviratne, Sachith and Suriyaarachchi, Hussel and Wen, Elliott and Nanayakkara, Suranga},
    pages = {3598--3602},
    volume = {2022-September},
    publisher = {International Speech Communication Association},
    doi = {10.21437/INTERSPEECH.2022-11258},
    issn = {19909772}
}

@article{kipf2016,
    title = {{Semi-Supervised Classification with Graph Convolutional Networks}},
    year = {2016},
    journal = {5th International Conference on Learning Representations, ICLR 2017 - Conference Track Proceedings},
    author = {Kipf, Thomas N. and Welling, Max},
    month = {9},
    publisher = {International Conference on Learning Representations, ICLR},
    url = {https://arxiv.org/abs/1609.02907v4},
    arxivId = {1609.02907}
}

@article{schuller2018,
    title = {{Speech emotion recognition}},
    year = {2018},
    journal = {Communications of the ACM},
    author = {Schuller, Björn W.},
    number = {5},
    month = {4},
    pages = {90--99},
    volume = {61},
    publisher = {ACMPUB27New York, NY, USA},
    url = {https://dl.acm.org/doi/10.1145/3129340},
    doi = {10.1145/3129340},
    issn = {15577317}
}

@article{Dissanayake2022SigRep:Learning,
    title = {{SigRep: Toward Robust Wearable Emotion Recognition with Contrastive Representation Learning}},
    year = {2022},
    journal = {IEEE Access},
    author = {Dissanayake, Vipula and Seneviratne, Sachith and Rana, Rajib and Wen, Elliott and Kaluarachchi, Tharindu and Nanayakkara, Suranga},
    pages = {18105--18120},
    volume = {10},
    publisher = {Institute of Electrical and Electronics Engineers Inc.},
    doi = {10.1109/ACCESS.2022.3149509},
    issn = {21693536}
}

@article{wu2023sslWearableEmo,
    title = {{Transformer-Based Self-Supervised Multimodal Representation Learning for Wearable Emotion Recognition}},
    year = {2023},
    journal = {IEEE Transactions on Affective Computing},
    author = {Wu, Yujin and Daoudi, Mohamed and Amad, Ali},
    publisher = {Institute of Electrical and Electronics Engineers Inc.},
    doi = {10.1109/TAFFC.2023.3263907},
    issn = {19493045},
    arxivId = {2303.17611}
}

@article{zhang2023sslreview,
  title={Self-Supervised Learning for Time Series Analysis: Taxonomy, Progress, and Prospects},
  author={Zhang, Kexin and Wen, Qingsong and Zhang, Chaoli and Cai, Rongyao and Jin, Ming and Liu, Yong and Zhang, James and Liang, Yuxuan and Pang, Guansong and Song, Dongjin and others},
  journal={arXiv preprint arXiv:2306.10125},
  year={2023}
}

@article{wu2023,
   author = {Yujin Wu and Mohamed Daoudi and Ali Amad},
   doi = {10.1109/TAFFC.2023.3263907},
   issn = {19493045},
   journal = {IEEE Transactions on Affective Computing},
   publisher = {Institute of Electrical and Electronics Engineers Inc.},
   title = {Transformer-Based Self-Supervised Multimodal Representation Learning for Wearable Emotion Recognition},
   year = {2023},
}

\vspace{12pt}

\end{document}